\PassOptionsToPackage{table}{xcolor}
\documentclass[10pt,twocolumn,letterpaper]{article}

\usepackage[pagenumbers]{cvpr} %
\usepackage[T1]{fontenc}
\usepackage{times}
\usepackage{amsmath}
\usepackage{amssymb}
\usepackage{graphicx}
\usepackage{subcaption}
\usepackage{wrapfig}
\usepackage{booktabs}
\usepackage{makecell}
\usepackage{multirow}
\usepackage{array}
\usepackage{tabularx}
\newcolumntype{C}[1]{>{\centering\arraybackslash}p{#1}}
\usepackage{algorithm}
\makeatletter
\let\labelindent\relax
\makeatother

\usepackage{algpseudocode}
\usepackage{enumitem}
\newlength{\labelindent}
\usepackage{multicol}
\usepackage{xurl}
\usepackage{comment}
\usepackage{xspace}
\definecolor{lightblue}{HTML}{E6F2FF}
\definecolor{lightpurple}{HTML}{F2E6FF}

\usepackage{microtype}

\definecolor{compassblue}{HTML}{3F78C0}
\newcommand{\ours}{\textup{\textsc{\textcolor{compassblue}{Compass}}}\xspace} %
\AtBeginDocument{%
  \crefname{table}{Table}{Tables}%
  }

\definecolor{cvprblue}{rgb}{0.21,0.49,0.74}
\usepackage[pagebackref,breaklinks,colorlinks,allcolors=cvprblue]{hyperref}

\def\confName{CVPR}
\def\confYear{2027}

\title{\ours: Bridging Agentic Reasoning and Dexterous Policies via \\ Compiled Task Fields}

\author{
Xiaoshen Han$^{1,2}$ \quad
Jiahang Cao$^{1,2}$ \quad
Xinghang Li$^{2}$ \quad
Tao Kong$^{2}$ \quad
Hang Lai$^{2\dagger}$ \quad
Weinan Zhang$^{1\dagger}$ \\[4pt]
$^{1}$Shanghai Jiao Tong University \quad
$^{2}$Xiaomi Robotics \\[2pt]
{\small $^{\dagger}$Corresponding authors}
}

\begin{document}
\makeatletter
\g@addto@macro\@maketitle{%
  \begin{minipage}{\textwidth}
    \centering
    \vspace{-24pt} %
    \includegraphics[width=\textwidth]{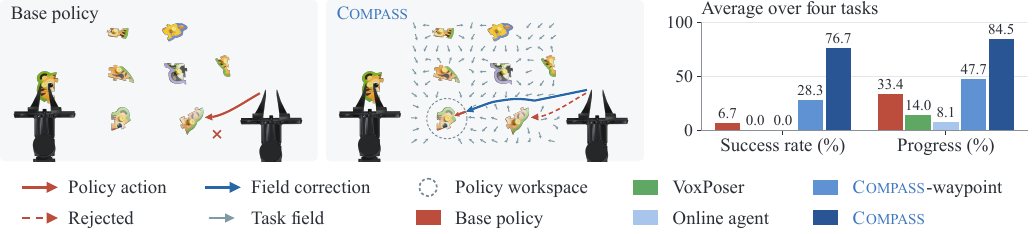}
    \captionof{figure}{\textbf{\ours.}
    Left: the base policy alone moves toward a wrong puzzle piece.
    Middle: in \ours, a VLM agent compiles a task field before execution.
    At run time, the field rejects action chunks that deviate from the agent's decisions and corrects them, whereas the policy keeps full control inside its workspace.
    Right: averaged over four tasks, \ours raises success from 6.7\% to 76.7\%.}
    \label{fig:teaser}
  \end{minipage}
  \vspace{4pt} %
}
\makeatother
\maketitle
\begin{abstract}
Everyday manipulation requires both reasoning and dexterity.
A robot must interpret vague instructions, distinguish targets from similar distractors,
keep track of progress, and plan spatial arrangements, while also
performing precise, coordinated actions.
Vision-language model (VLM) agents handle the reasoning well, but they are too
slow and coarse for real-time control, whereas visuomotor policies
show the opposite strengths.
We present \ours (COMpiled Potentials for Action Selection and Steering), which bridges agentic reasoning and dexterous control through compiled task fields.
Before execution, the agent reasons about the instruction and the
scene, compiling its decisions into a stateful potential field over the policy's action space.
At run time, the field evaluates each action chunk proposed by a frozen visuomotor policy.
It keeps chunks that make progress, corrects those that deviate from the agent's decisions, and leaves the policy in full control near each target.
With no agent queries in the control loop, execution remains smooth while preserving the policy’s reactivity and robustness during contact-rich manipulation.
On four real-world bimanual tasks, \ours improves the average success rate from
\textbf{6.7\%} to \textbf{76.7\%}.
Further experiments highlight the complementary roles of agent reasoning and policy motor skills.
Project page: \url{https://compass-field.github.io}.
\end{abstract}

\section{Introduction}
\label{sec:intro}

Everyday manipulation requires both high-level reasoning and precise, adaptive motor control.
Tidying a desk, for example, requires deciding which items to keep or discard, remembering what has already been put away, and planning how to arrange the remaining items in a drawer~\cite{wu2023tidybot}.
At the same time, it requires precisely grasping small objects and carefully aligning them for placement into fitted compartments.
A robot cannot complete such tasks if either capability is missing, just as patients with frontal-lobe damage fail to organize multi-step activities~\cite{shallice1991deficits,schwartz1991quantitative} and patients with cerebellar lesions move inaccurately~\cite{manto2012cerebellum,bastian1996cerebellar}.

Frontier large vision-language models (VLMs) and agentic systems built on them now meet much of the first demand. 
They follow open-ended instructions~\cite{ahn2022saycan,huang2022inner}, distinguish similar objects~\cite{yang2023setofmark,you2024ferret}, track task progress~\cite{du2023success,ma2025gvl}, and reason about spatial layouts~\cite{huang2023voxposer,huang2024rekep,chen2024spatialvlm,yuan2024robopoint}.
Translating these capabilities into dexterous motor control, however, remains challenging. Recent works use VLMs to control robot arms through direct commands, generated programs, or cost functions for motion planning~\cite{showharness,agp,gptpolicy,liang2023code,huang2023voxposer,huang2024rekep,pan2025omnimanip}. These systems support a range of manipulation tasks, but still struggle with contact-rich tasks requiring precise closed-loop control~\cite{robocurve,pantograph}.

By contrast, visuomotor policies trained from demonstrations excel at precise grasping, insertion, and coordinated bimanual manipulation at high control rates~\cite{zhao2023act,chi2023diffusion,black2024pi0,black2025pi05,liu2026oat,liu2026orderedtokens,han2026re3sim}. Recent benchmarks, however, reveal their difficulty with subtle instruction changes~\cite{shi2025hirobot,zhou2025liberopro,fei2025liberoplus}, reasoning about spatial relationships~\cite{lin2025evo0,li2025qdepthvla}, and maintaining performance among distractors~\cite{wang2025vlatest}. Limited access to extended observation histories further leaves many policies unable to reliably track completed steps, undermining performance on
tasks whose next action depends on earlier
events~\cite{shi2026memoryvla,fang2025sam2act}.

A natural direction is to combine
agentic reasoning with the motor skills of learned policies. One line of work uses a high-level planner or VLM to decompose a task into language subgoals that prompt a low-level policy~\cite{ahn2022saycan,shi2025hirobot}. However, language alone is a brittle interface: even policies trained on large datasets often fail to ground diverse instructions~\cite{emukpere2026robust,hou2026langgap,glossop2025cast,fei2025liberoplus}, and such prompts provide no explicit mechanism for detecting and correcting action deviations.
Another line places an agent or VLM in the execution loop to monitor and correct the policy~\cite{fpcvla,volo,wu2025foresight,astra_embodied}.
Querying a large model throughout execution is slow and costly, limiting
the practicality of such systems for real-time manipulation. These limitations motivate an interface that carries high-level decisions
into closed-loop execution without repeated VLM queries or agent tool calls, while preserving
the policy's local adaptability and robustness during contact.

To this end, we present \ours (COMpiled Potentials for Action Selection and Steering), 
an approach built on the observation that many
high-level decisions depend only on the instruction and the initial scene, and can therefore be made before execution. A VLM agent makes these decisions and compiles them into a
\emph{task field}, a scalar potential over the policy's action space paired with a
compact discrete state that tracks stages and counts. This state is updated online from measured feedback without further VLM queries.
At run time, the field checks each action chunk proposed by the policy,
keeping those that make progress and correcting those that deviate from
the agent's decisions.
Near the target, it leaves the policy in full control of contact-rich manipulation.

We evaluate \ours on four real-world bimanual tasks covering instruction following, robustness to similar distractors, long-horizon
memory, and spatial planning.
Across these tasks, \ours raises the average success rate by 70.0 percentage points over the frozen base policies, while consistently outperforming a planner-based method, an online agent, and a waypoint-based
handoff variant. Ablations across six agent models and four visuomotor policies further highlight the importance of both strong high-level reasoning and capable low-level motor skills.

Our contributions are threefold:
\begin{itemize}[leftmargin=*]
  \item We propose \ours, which uses compiled task fields as an interface
between agentic reasoning and frozen visuomotor policies.
The interface carries task decisions and memory into closed-loop
execution without online VLM queries, while preserving the policy's
local adaptability during contact-rich manipulation.
  \item We conduct over 500 real-robot trials on four bimanual tasks with
  six agent models and four policies. \ours raises the average success
  rate from 6.7\% to 76.7\% and assembles a jigsaw puzzle among
  distractor pieces in 11/15 trials, whereas no baseline exceeds 2/15.
  \item By comparing agent models and policies,
we characterize the complementary roles of reasoning and motor skill
in everyday manipulation.
Strong task reasoning cannot compensate for inadequate contact skills,
while precise motor execution alone does not ensure correct task
completion.
\end{itemize}

\section{Related Work}
\label{sec:related}

\noindent\textbf{Agentic and VLM-driven planning and control.}
Large language and vision-language models are widely used to plan and control robot tasks. Early systems use language models to select pretrained skills~\cite{ahn2022saycan}, refine plans with language feedback~\cite{huang2022inner}, and infer object-placement preferences from user examples~\cite{wu2023tidybot}. Code as Policies instead  writes programs that call perception and control APIs~\cite{liang2023code}.
Another line translates language into spatial objectives such as 3D value maps~\cite{huang2023voxposer}, keypoint constraints~\cite{huang2024rekep}, affordance points or part-level constraints~\cite{liu2024moka,huang2024copa}, and interaction primitives~\cite{pan2025omnimanip}.
These approaches rely on motion planners or controllers to execute the motions, even for contact-rich manipulation. 
Recent agentic systems also let frontier VLMs act directly during execution through end-effector commands, tool calls, or generated programs~\cite{showharness,agp,gptpolicy,astra_embodied}. However, recent evaluations show difficulties in precise,
contact-rich manipulation~\cite{robocurve,pantograph},
while repeated model queries introduce execution latency~\cite{pantograph,astra_embodied}. Our task field is likewise an agent-compiled spatial objective, built on artificial potential
fields~\cite{khatib1986potential,rimon1992navigation}, but it guides a learned policy instead of generating the contact motion, and keeps agent queries out of the control loop.

\noindent\textbf{Learned visuomotor policies.}
Contact is where policies learned from demonstrations excel.
Task-specific policies model diverse manipulation
behaviors~\cite{florence2021ibc,shafiullah2022bet,zhao2023act,chi2023diffusion,ze2024dp3}
and reach high precision on bimanual
tasks~\cite{zhao2023act,fu2024mobilealoha}.
Language conditioning extends learned policies to multi-task settings~\cite{jang2022bcz,shridhar2022peract,brohan2023rt1,team2024octo},
and vision-language-action (VLA) models scale further by combining pretrained VLMs with large robot
datasets~\cite{brohan2023rt2,kim2024openvla,black2024pi0,black2025pi05,xiaomiroboticsteam2026xiaomirobotics1scalingvisionlanguageactionmodels}.
Recent work improves inference and execution efficiency for VLA models, enabling smooth,
real-time control~\cite{black2025rtc,han2026bspline}.
Even with a VLM backbone, VLA policies remain sensitive to
out-of-distribution instructions~\cite{shi2025hirobot,fei2025liberoplus}
and similar distractors~\cite{wang2025vlatest},
while struggling with spatial reasoning~\cite{lin2025evo0}
and long-horizon memory~\cite{shi2026memoryvla}.
\ours addresses these limitations by supplying the missing guidance through the compiled task field, without further policy training.

\begin{figure*}[t]
  \centering
  \includegraphics[width=\linewidth]{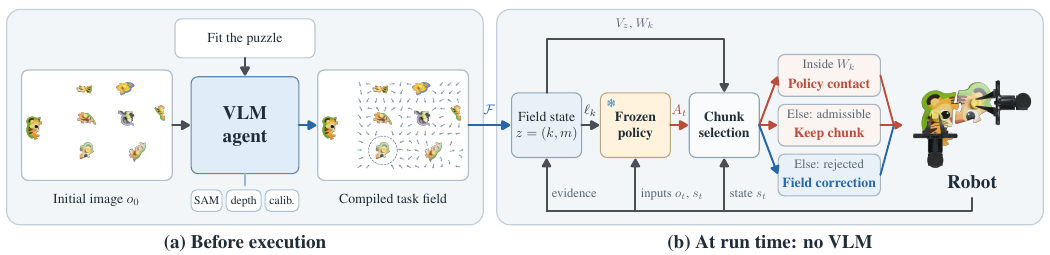}
  \caption{\textbf{Overview of \ours.}
  (a) Before execution, the agent compiles the instruction and the
  initial image into a task field, which specifies for each stage the
  target to approach and the non-targets to avoid.
  (b) At run time, chunk selection checks each chunk of the frozen policy,
  which is then executed, replaced by a field correction, or executed
  freely inside the policy workspace $W_k$.
  Measured feedback updates the field state $z=(k,m)$, and no agent is
  queried in this loop.}
  \label{fig:overview}
  \vspace{-5pt}
\end{figure*}

\noindent\textbf{Bridging reasoning and low-level policies.}
Existing approaches connect high-level reasoning to low-level policies
through different interfaces.
Hierarchical systems condition policies on intermediate representations
such as language subgoals, 2D paths, object masks, or trajectory
sketches~\cite{shi2025hirobot,belkhale2024rth,li2025hamster,huang2025roboground,gu2023rttrajectory,steerable}.
These interfaces require policies trained to interpret the chosen
representation, limiting compatibility with pretrained policies that
lack such conditioning.
Online monitoring systems instead keep a model in the execution loop
to select, correct, or re-prompt the
policy~\cite{fpcvla,criticloop,robotcritics,stageaware,volo,wu2025foresight,astra_embodied},
introducing additional online model queries.
Another line combines or steers behaviors at inference time.
Compositional methods combine learned distributions, policies, skills,
or behaviors~\cite{liu2022composable,ajay2023decisiondiffuser,wang2024poco,cao2026gpc,mishra2023gsc,chen2026costream},
while steering methods use learned value functions, dynamics models,
or human input to guide frozen
policies~\cite{nakamoto2024vgps,du2025dynaguide,wang2025itps}.
SkipVLA replaces free-space VLA steps with a classical
planner~\cite{agrawal2026skipvla}, resembling our \ours-waypoint variant.
Others guide a frozen policy with spatial scores built by a VLM, either
through the gradients of its sampler~\cite{liu2026vls} or by re-ranking
its proposals once it stalls~\cite{xu2025afi}. 
\ours instead targets semantic decisions that the policy cannot make, such
as which of several similar instances to take or how many remain.
Its field can replace a chunk entirely, and its stages advance
from measured evidence without a VLM.

\providecommand{\SO}{\mathrm{SO}(3)}

\section{Method}
\label{sec:method}

As \cref{fig:overview} shows, in \ours, a VLM agent guides a policy through a \emph{task field}, a
stateful potential over the policy's action space
(\cref{sec:formulation,sec:field}).
Before execution, the agent reasons about the instruction and the
initial scene once, compiling its high-level decisions into the field
(\cref{sec:compile}).
At run time, the field checks each action chunk proposed by the policy,
corrects chunks that deviate from the agent's decisions, and leaves
contact-rich motion to the policy without querying the VLM
(\cref{sec:execution}).

\subsection{Problem Formulation}
\label{sec:formulation}

\noindent\textbf{Setting.}
We consider a bimanual robot that carries out a natural-language
instruction $L$ under a policy $\pi$.
Each arm is commanded by an absolute target action
$a = (p, R, g) \in \mathcal{A} = \mathbb{R}^3 \times \SO \times \mathbb{R}$,
consisting of the gripper-tip position $p$ (the midpoint between the two
fingertips), the gripper orientation $R$, and the gripper opening $g$.
The measured arm state $s_t \in \mathcal{A}$ shares this form, and we
omit arm indices.
At step $t$, the policy receives the measured state $s_t$, a short prompt $\ell$, and an
observation $o_t$ comprising images from one head camera and two wrist
cameras and predicts $H$ actions relative to $s_t$,
which compose with $s_t$ into a chunk of absolute actions
$A_t = s_t \oplus \pi(s_t, o_t, \ell) = (a_{t+1}, \dots, a_{t+H}) \in \mathcal{A}^H$.

\noindent\textbf{Division of labor.}
\ours separates high-level task reasoning from low-level motor control.
High-level decisions interpret the instruction and the scene to
determine task stages, target identities, and spatial goals, whereas
low-level decisions govern physical contact and continuous motion.
Before execution, the VLM agent resolves the high-level decisions from
$L$ and $o_0$, compiling them into the task field.
At run time, the policy proposes action chunks throughout the task.
Away from the target, the field checks each chunk across position,
orientation, and gripper opening, intervening only when the policy
deviates from the agent's decisions.
Near the target, the field leaves the policy in full control of every
degree of freedom to handle contact-rich manipulation.

\subsection{Task Field Representation}
\label{sec:field}

The task field encodes the agent's high-level decisions as a potential
over actions and a discrete state of task progress.

\noindent\textbf{Definition.}
Formally, the agent compiles a task field
\begin{equation}
  \mathcal{F} = \big( V,\; \{W_k, \ell_k, E_k\}_{k=1}^{K},\; z_0,\; \delta \big) .
  \label{eq:taskfield}
\end{equation}
Here $V : \mathcal{A} \times \mathcal{Z} \to \mathbb{R}_{\ge 0}$ is a
scalar potential that scores an action $a$ under a discrete field state
$z = (k, m) \in \mathcal{Z}$, abbreviated as $V_z(a) = V(a; z)$.
The state $z$ consists of the current stage $k \in \{1, \dots, K\}$ and
a memory ledger $m$.
Each stage $k$ represents one step of the task, such as approaching or
placing an object, and specifies a policy workspace $W_k$, a policy
prompt $\ell_k$, and transition predicates $E_k$.
Starting from $z_0$, the transition function $\delta$ updates $z$ from
measured feedback.
The stage-specific parameters of $V$ are
described in \cref{sec:compile}.
All components are evaluated from robot states and camera
images at control rate without VLM queries.

\noindent\textbf{Potential terms.}
The potential $V_z$ assigns low values to actions that satisfy the
stage's semantic goals and high values to actions that violate its
constraints, such as approaching distractors or obstacles.
For an action $a = (p, R, g)$, it  combines up to five terms (parameters in
\cref{sec:supp-field}):
\begin{equation}
  \begin{aligned}
  V_z(a) ={}& \underbrace{\tfrac12 \big\| \Sigma_k^{-1}(p - p_k^\ast) \big\|^2}_{\text{target}}
  + \underbrace{\textstyle\sum_{j} \tfrac{w_j}{2} \big[\tfrac{r_j - d_j(p)}{r_j}\big]_+^2}_{\text{clearance}} \\
  &+ \underbrace{\textstyle\sum_{i \in \mathcal{N}_z} \phi_i(p)}_{\text{non-target}}
  + \underbrace{\lambda_R\, \psi(R, R_k^\ast)}_{\text{orientation}}
  + \underbrace{\lambda_g \big(g - g_k^\ast(p)\big)^2}_{\text{gripper}} .
  \end{aligned}
  \label{eq:field}
\end{equation}
The target term pulls the gripper tip toward the stage goal $p_k^\ast$
with tolerance $\Sigma_k$.
The clearance term penalizes a signed distance $d_j$ smaller than the
margin $r_j$ to each obstacle $j$, such as the table, the container
walls, or objects that are already packed.
The non-target term places a bump $\phi_i$ on every region
$i \in \mathcal{N}_z$ that the arm must avoid, such as a similar
distractor.
Each bump is a smooth function fitted to the 3D extent of its region,
with finite support, so it acts only near that region
(\cref{sec:supp-field}).
The orientation term $\psi$ is the geodesic angle between $R$ and
the desired stage orientation $R_k^\ast$, or the angle between their approach axes when twist about this axis is free.
The gripper term keeps the opening close to a target $g_k^\ast(p)$,
which stays open during the approach, remains closed while an object is carried,
and transitions smoothly to open near the placement goal.

\noindent\textbf{Choice of a scalar potential.}
We use a scalar potential to evaluate policy actions and generate
corrections by gradient descent (\cref{sec:execution}).
Its terms are additive and in closed form, so $V_z$ and its gradient
are cheap to evaluate on the CPU. 
For a bimanual robot, each arm has its own potential in every stage. We do not sum the
two because coordination between the arms is a low-level decision that
remains with the policy.

\noindent\textbf{Field state.}
The field state holds all task memory, so the policy does not need to
track progress.
The ledger $m$ records the remaining targets and their identities, the
quota and count of each category, and the retry budgets.
The non-target set $\mathcal{N}_z$ is recomputed from the ledger, so
memory changes the potential rather than the policy. For example, each confirmed release increments the corresponding category count, and once its quota is met, the remaining instances of that category join $\mathcal{N}_z$.

\noindent\textbf{Stage transitions.}
The field state advances only on measured evidence.
Each predicate in $E_k$ returns \emph{true}, \emph{false}, or
\emph{unknown} based on evidence such as gripper closure, lift, release near the placement goal,
and the appearance of the target in the wrist images.
The transition function $\delta$ advances the stage only when the required predicates return \emph{true},
so missing evidence is never treated as success.
A detected failure, such as an empty grasp or grasping a wrong object, leads to a
recovery stage with its own retry budget (\cref{sec:supp-field,sec:supp-exec}).

\subsection{Task Field Compilation}
\label{sec:compile}

\noindent\textbf{Robot and task knowledge.}
Compiling a 3D field from an image requires both a camera-to-robot
transformation and knowledge of the task structure.
Once per robot, we estimate this transformation from a few annotated gripper landmarks across reachable poses without a calibration board, simplifying deployment
(\cref{sec:supp-calib}).
Once per task, the agent builds a reusable \emph{skill}
document~\cite{rax,astra_embodied} by reading the task description,
inspecting a few demonstrations, and briefly exploring the task on
the robot (\cref{sec:supp-agent}).
The skill specifies the stage templates and transition criteria shared
across all layouts of the task, whereas the targets, spatial goals, and
potential terms are compiled afresh for each layout.

\noindent\textbf{Compilation steps.}
For each layout, the robot starts at its home pose, and the agent receives
the instruction $L$ and the initial head-camera image in $o_0$.
Using a promptable segmenter~\cite{kirillov2023segment,carion2025sam3}, a monocular depth model~\cite{lin2025da3}, 
a library of potential terms, and a renderer, the agent compiles the field in four steps without privileged object state.
First, the agent parses the instruction into ordered stages and assigns
each stage a prompt $\ell_k$ in the style of the policy's training
prompts.
Second, the agent marks relevant objects, object parts, and containers in the image with point prompts.
The segmenter produces 2D masks, which are lifted into 3D bounding
extents in the robot frame using the estimated depth.
These regions, together with the stage order and category quotas,
initialize the ledger $m_0$ and the non-target sets $\mathcal{N}_z$.
Third, the agent computes the spatial goal of each stage, such as a grasp pose, placement slot, or assembly orientation, setting the target pose $(p_k^\ast, R_k^\ast)$ and gripper schedule $g_k^\ast(p)$.
Finally, it assembles the potential terms, policy workspaces $W_k$, and
transition predicates $E_k$ for all stages, projects the compiled field geometry onto the head image for
visual verification, and revises any misaligned terms (\cref{sec:supp-agent}).

\noindent\textbf{Selective term activation.}
Adding unnecessary terms to the field would interfere with motions that
the policy already performs well.
The agent therefore includes a potential term only when it encodes task
requirements that the policy may not reliably satisfy on its own.
For each included term, the agent specifies the stages in which it is
active and the transition that deactivates it, keeping the field
sparse and leaving purely contact-rich stages unconstrained.
For example, a bump on a distractor is justified by the requested
target identity, acts only around that distractor, and is removed once
the target is grasped.

\begin{figure*}[t]
  \centering
  \begin{subfigure}[t]{0.245\linewidth}
    \centering
    \includegraphics[width=\linewidth]{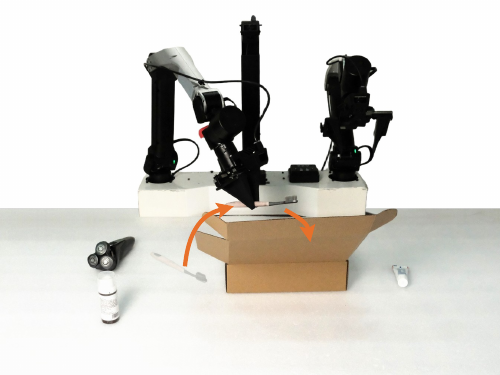}
    \caption{Pack box}
    \label{fig:tasks-illus-box}
  \end{subfigure}\hfill
  \begin{subfigure}[t]{0.245\linewidth}
    \centering
    \includegraphics[width=\linewidth]{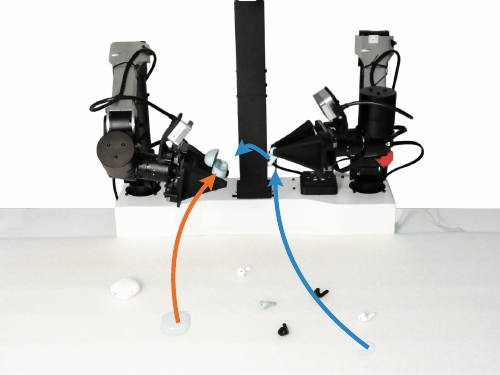}
    \caption{Pack earphone}
    \label{fig:tasks-illus-earphone}
  \end{subfigure}\hfill
  \begin{subfigure}[t]{0.245\linewidth}
    \centering
    \includegraphics[width=\linewidth]{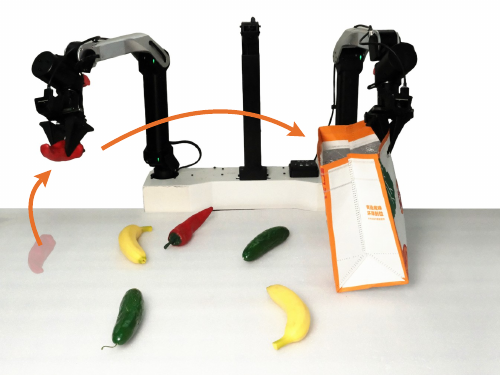}
    \caption{Count fruit}
    \label{fig:tasks-illus-fruit}
  \end{subfigure}\hfill
  \begin{subfigure}[t]{0.245\linewidth}
    \centering
    \includegraphics[width=\linewidth]{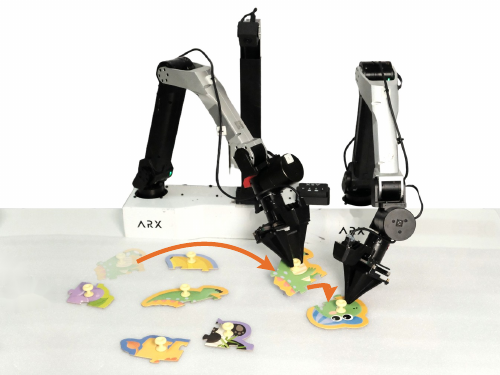}
    \caption{Puzzle}
    \label{fig:tasks-illus-puzzle}
  \end{subfigure}

  \caption{\textbf{Real-robot tasks.}
  One test layout of each task during execution.
  Moved objects are shown at their initial positions at 30\% opacity, and orange and blue arrows indicate the left and right arms' motions in the camera view.
  \Cref{sec:supp-tasks} shows all 15 test layouts of each task.}
  \label{fig:tasks}
  \vspace{-5pt}
\end{figure*}

\subsection{Field-Guided Execution}
\label{sec:execution}

Once compiled, $\mathcal{F}$ governs closed-loop execution over successive action chunks $A_t$, updating the field state $z=(k,m)$ via $\delta$ after each executed chunk (\cref{alg:execution}).

\noindent\textbf{Chunk selection.}
Outside the policy workspace $W_k$, we evaluate each proposed chunk $A_t$ along a densely interpolated path $\tilde{a}_{0:N}$ from $s_t$ through $A_t$.
Let $\Phi_z(a) = \sum_{i \in \mathcal{N}_z} \phi_i(p)$ denote the non-target potential at $a=(p,R,g)$.
We keep $A_t$ if
\begin{equation}
  \begin{aligned}
  &V_z(a_{t+H}) - V_z(s_t) \le \tau, \\
  &\max_n \Phi_z(\tilde{a}_n) \le \max\big(\epsilon, \Phi_z(s_t)\big), \\
  &\Phi_z(a_{t+H}) < \Phi_z(s_t) \ \text{ if } \ \Phi_z(s_t) > \epsilon ,
  \end{aligned}
  \label{eq:admissible}
\end{equation}
where $\tau$ is a small tolerance on potential increase and $\epsilon$ is the threshold below which a bump is considered inactive.
The first condition checks the net potential change only between the start and end of the chunk, leaving the policy free to choose its intermediate path.
The remaining two prevent the path from entering non-target regions, or if the arm already starts inside one ($\Phi_z(s_t) > \epsilon$), require the chunk to move it out without going deeper.

\noindent\textbf{Field correction.}
A rejected chunk is replaced by $H$ actions generated by bounded gradient descent on $V_z$.
Starting at $s_t$, each step maps the current action $a = (p, R, g)$ to
the next action $a' = (p', R', g')$:
\begin{equation}
  \begin{aligned}
  p' &= p - \eta_p \nabla_p V_z(a), \qquad g' = g - \eta_g\, \partial_g V_z(a), \\
  R' &= \exp\!\big(-\eta_R [\nabla_R V_z(a)]_\times\big)\, R ,
  \end{aligned}
  \label{eq:descent}
\end{equation}
where $\nabla_R V_z(a) \in \mathbb{R}^3$ is the rotation gradient on $\SO$
and $[\cdot]_\times$ is the skew-symmetric operator.
At each step, we backtrack the step sizes $(\eta_p, \eta_R, \eta_g)$
together until $V_z$ decreases and the step avoids non-target regions.
Across the $H$ steps, the cumulative correction is bounded by
per-chunk budgets.

\noindent\textbf{Detour around local minima.}
Gradient descent can stall in a local minimum when a non-target bump lies between the current position $p_t$ and the stage goal $p_k^\ast$.
Because the potential is in closed form, we also evaluate it along the straight line from $p_t$ to $p_k^\ast$.
If this line crosses a bump or descent stalls early, we insert a temporary goal outside the support of all blocking bumps before proceeding to $p_k^\ast$.

\noindent\textbf{Policy workspace.}
Each stage defines a workspace $W_k \subset \mathcal{A}$ around its
target, sized by the agent to cover expected perception error.
Once the \emph{measured} state $s_t$ enters $W_k$, the policy controls
every degree of freedom, while the field only checks that its actions
remain within $W_k$.
Small boundary violations are repaired by a bounded position shift, and larger ones trigger a new policy query. Further execution details are given in \cref{sec:supp-exec}.

\section{Experiments}
\label{sec:exp}

We evaluate \ours on four real-world bimanual tasks to answer four questions.
\textbf{Q1}: Which capabilities does the task field add to a frozen policy
(\cref{sec:main-results})?
\textbf{Q2}: How does \ours compare in success and speed with other paradigms for bridging agentic reasoning and motor control (\cref{sec:comparison})?
\textbf{Q3}: Does success require both a capable agent and a capable policy
(\cref{sec:models,sec:policies})?
\textbf{Q4}: Given the same high-level plan, which interface best guides the policy
(\cref{sec:interface})?\looseness=-1

\subsection{Setup}
\label{sec:setup}

\noindent\textbf{Robot and policies.}
We evaluate on bimanual robots with two six-DoF ARX X5 arms, parallel grippers, one head camera, and two wrist cameras (\cref{fig:tasks}).
We use XR1~\cite{xiaomiroboticsteam2026xiaomirobotics1scalingvisionlanguageactionmodels} for pack earphone and puzzle, which require precise insertion and benefit from its pretraining on the same embodiment, and $\pi_{0.5}$~\cite{black2025pi05} for pack box and count fruit.
Each policy is post-trained on task demonstrations (\cref{sec:impl}), predicts chunks of $H = 30$ actions (1\,s at 30\,Hz), and is kept frozen in all experiments.\looseness=-1

\begin{figure*}[t]
  \centering
  \includegraphics[width=\linewidth]{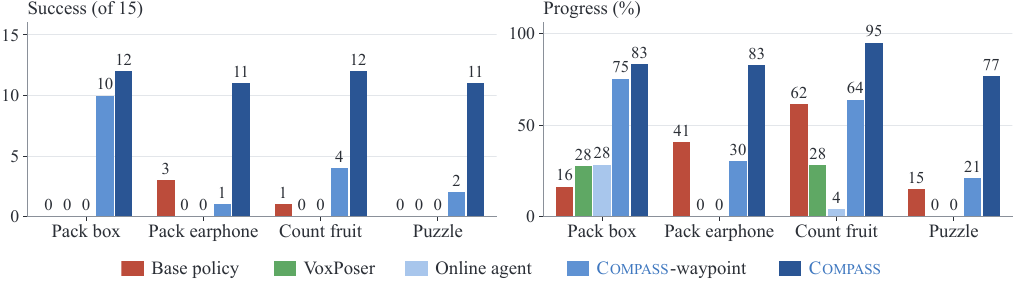}
  \caption{\textbf{Main results.}
  Left: number of successful trials out of 15. Right: average task progress across the four tasks.
  \ours outperforms all baselines on every task by combining task-level reasoning with precise contact.}
  \label{fig:main}
  \vspace{-5pt}
\end{figure*}

\noindent\textbf{Agent and perception.}
By default, the VLM agent uses GPT-6-Astra with medium reasoning effort in the Codex CLI, paired with SAM~3~\cite{carion2025sam3} and Depth Anything~3~\cite{lin2025da3} for perception (\cref{sec:supp-compile}).\looseness=-1

\noindent\textbf{Tasks.}
Each task mainly focuses on one capability (\cref{fig:tasks}).
\emph{Pack box} tests instruction following: the robot packs 3--6 objects
into a box in the order named by a free-form request.
\emph{Pack earphone} tests robustness to similar distractors: the robot
packs 2--3 earphone sets, some of which look nearly identical,
each into its own case.
\emph{Count fruit} tests long-horizon memory: one arm holds a deep bag open
while the other packs fruit and vegetables according to quotas such as
``a couple of each,'' with packed items hidden inside the bag from all three cameras.
\emph{Puzzle} tests spatial planning: the robot assembles a jigsaw puzzle
among distractor pieces from other puzzles.
For each task, we evaluate on 15 random test layouts spanning both seen and unseen target combinations, with cross-station tabletop variations (\cref{sec:supp-tasks}).\looseness=-1

\noindent\textbf{Baselines.}
We compare \ours with three baselines and a waypoint-handoff variant (\ours-waypoint).
All methods share the same robot, perception models, and VLM when applicable (\cref{sec:supp-baselines}).\looseness=-1
\begin{itemize}[leftmargin=*,nosep]
  \item \emph{Base policy} runs the frozen policy without the agent, using the task's training-style prompt rather than the user instruction.\looseness=-1
  \item \emph{VoxPoser}~\cite{huang2023voxposer} uses no learned policy: the VLM composes 3D affordance and avoidance value maps from the same perception, and a motion planner generates all end-effector motion along these maps.\looseness=-1

  \item \emph{Online agent} adapts the hybrid mode of~\cite{astra_embodied} to our robot: GPT-6-Astra is queried online at every chunk to either execute a prefix of the policy chunk or replace it with end-effector waypoints.
  It shares the initial perception, stage ledger, and transitions with \ours, so
  only the field is replaced.\looseness=-1
  \item \emph{\ours-waypoint} also shares the compiled stages, targets, and transitions with \ours: a planner moves the arm to a pre-grasp pose above each target and hands control to the policy within 2.5\,cm, so this variant isolates the effect of the field.\looseness=-1
\end{itemize}

\noindent\textbf{Metrics.}
All methods are evaluated on the same 15 layouts of each task, with one
trial per layout.
Each sub-goal corresponds to one packed target or fitted piece, with partial credit for intermediate steps (\cref{sec:supp-eval}).
A trial succeeds if all sub-goals are completed, and task progress is the
fraction of sub-goals completed, averaged over the 15 trials.
Each trial stops after 20\,min or at an unrecoverable error (e.g., packing a wrong, out-of-order, or excess item).\looseness=-1

\begin{figure*}[t]
  \centering
  \includegraphics[width=\linewidth]{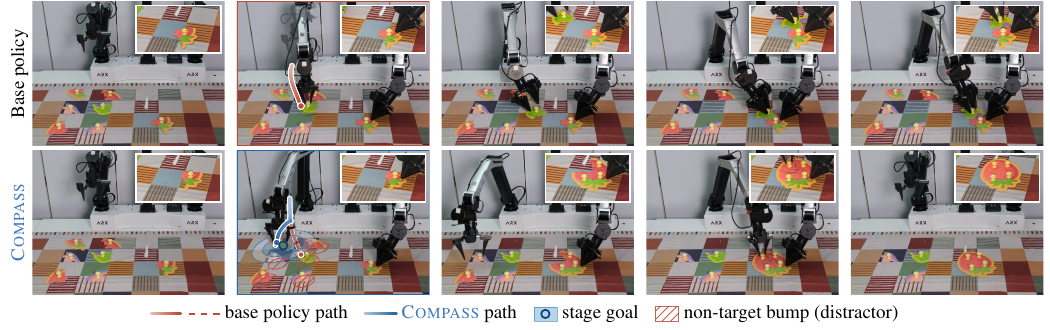}
  \caption{\textbf{Qualitative comparison.}
  Base policy and \ours on the same puzzle layout.
  Each row shows one trial from left to right, with a time-aligned close-up in the top-right inset and the final state in the last column.
  In the second column, the red line is the base policy path toward a distractor piece, while the blue line is the \ours path toward the stage goal on the target piece. The red regions are non-target bumps on distractor pieces.}
  \label{fig:qualitative}
  \vspace{-5pt}
\end{figure*}

\subsection{Effect of the Task Field}
\label{sec:main-results}

\Cref{fig:main} shows that \ours achieves the highest success and progress
on all four tasks.
It succeeds in 11/15 or 12/15 trials on every task, whereas no other
method exceeds 4/15 on pack earphone, count fruit, or puzzle, since these
tasks demand both task-level reasoning and precise contact.
The base policy performs the local skills, achieving 15.0--61.5\%
task progress, yet it succeeds in at most 3/15 trials on any task.
In answer to \textbf{Q1}, each task exposes one capability that the task field adds to the same frozen policy:\looseness=-1
\begin{itemize}[leftmargin=*,nosep]
  \item \emph{Instruction following.}
  On pack box, the base policy rarely fails to grasp or place an object,
  but it packs the objects out of the requested order or skips some of
  them, and its progress falls from 34.7\% on layouts with three or four
  objects to 4.1\% on layouts with five or six.
  \ours raises success from 0/15 to 12/15, since the agent maps each
  requested object to its instance and to a name from the policy's
  training prompts, and the field guides the arm through them in order.\looseness=-1
  \item \emph{Robustness to similar distractors.}
  On pack earphone, the base policy often places an earbud into the wrong case.
  \ours raises success from 3/15 to 11/15: the ledger records which
  earbuds belong to each set, and the field rejects 62 chunks across the 15 trials that head
  toward a same-color earbud of another set.\looseness=-1
  \item \emph{Long-horizon memory.}
  On count fruit, the base policy reaches 61.5\% task progress but
  succeeds in only 1/15 trials, since packed items are hidden inside the bag and the policy sees only the current frame.
  With the ledger keeping count, \ours succeeds in 12/15 trials, and its
  three failed trials still average 76\% task progress.\looseness=-1
  \item \emph{Spatial planning.}
  On puzzle, the base policy fails in all 15 trials, mostly because it
  joins pieces that do not fit together.
  \ours succeeds in 11/15 trials, since the agent decides which piece
  joins the anchor at each stage and the field keeps the arm away from
  the distractor pieces.\looseness=-1
\end{itemize}
We also show a qualitative comparison on the puzzle task in \cref{fig:qualitative}: the base policy mistakes a distractor grape piece for the target and grabs it at the very first step, whereas \ours, guided by the task field, selects and fits the correct piece at every stage to complete the puzzle. Side-by-side rollout videos on all four tasks are provided in the supplementary material.\looseness=-1

\begin{table*}[t]
  \centering
  \vspace{3pt} %
    \caption{\textbf{Execution time comparison.}
    Results on pack box, pack earphone, and count fruit.
    SR is success out of 15 trials, Prog.\ is task progress in \%, Time / sub-goal is mean execution time in seconds per completed sub-goal over trials with nonzero progress, and Time / trial is mean execution time in seconds per trial.
    Times are mean\,$\pm$\,std, and -- indicates no completed sub-goal.
    Puzzle is omitted because all baselines complete too few pieces for a meaningful time comparison.}
  \label{tab:time}
  \setlength{\tabcolsep}{4pt} %
  \resizebox{\textwidth}{!}{%
  \begin{tabular}{l cccc cccc cccc}
    \toprule
    & \multicolumn{4}{c}{Pack box} & \multicolumn{4}{c}{Pack earphone} & \multicolumn{4}{c}{Count fruit} \\
    \cmidrule(lr){2-5}\cmidrule(lr){6-9}\cmidrule(lr){10-13}
    Method & SR & Prog. & Time / sub-goal & Time / trial & SR & Prog. & Time / sub-goal & Time / trial & SR & Prog. & Time / sub-goal & Time / trial \\
    \midrule
    Base policy    & 0/15 & 16.3 & 19.4\,$\pm$\,14.8 & 23.0\,$\pm$\,37.4 & 3/15 & 40.6 & 208.9\,$\pm$\,85.2 & 176.3\,$\pm$\,94.3 & 1/15 & 61.5 & 18.3\,$\pm$\,14.8 & 74.8\,$\pm$\,36.2 \\
    VoxPoser       & 0/15 & 27.6 & 380.0\,$\pm$\,167.8 & 465.8\,$\pm$\,207.4 & 0/15 & 0.0 & -- & 1200.0\,$\pm$\,0.0 & 0/15 & 28.3 & 197.7\,$\pm$\,83.7 & 458.3\,$\pm$\,209.2 \\
    Online agent   & 0/15 & 28.4 & 863.6\,$\pm$\,377.2 & 1200.0\,$\pm$\,0.0 & 0/15 & 0.0 & -- & 1200.0\,$\pm$\,0.0 & 0/15 & 4.0 & 1733.3\,$\pm$\,679.9 & 1200.0\,$\pm$\,0.0 \\
    \ours-waypoint & 10/15 & 75.4 & 20.1\,$\pm$\,11.9 & 65.2\,$\pm$\,23.1 & 1/15 & 30.3 & 350.1\,$\pm$\,154.8 & 228.4\,$\pm$\,120.9 & 4/15 & 63.9 & 13.7\,$\pm$\,5.4 & 111.6\,$\pm$\,50.5 \\
    \ours          & 12/15 & 83.3 & 38.1\,$\pm$\,29.9 & 154.0\,$\pm$\,51.4 & 11/15 & 82.8 & 376.1\,$\pm$\,226.9 & 598.9\,$\pm$\,212.5 & 12/15 & 95.2 & 17.7\,$\pm$\,11.0 & 119.6\,$\pm$\,36.7 \\
    \bottomrule
  \end{tabular}}
  \vspace{-5pt}
\end{table*}

\subsection{Comparison with Other Methods}
\label{sec:comparison}

To answer \textbf{Q2}, we compare \ours with VoxPoser, the online agent, and
\ours-waypoint (\cref{fig:main,tab:time}).\looseness=-1

\noindent\textbf{Fine manipulation still needs a learned policy.}
VoxPoser, which plans all motion from value maps composed by the VLM,
fails every trial on all four tasks.
It completes no sub-goal on pack earphone or puzzle, where its planned
motion is not precise enough to insert an earbud or fit a piece.
\ours instead leaves this contact to the policy inside each
workspace $W_k$.\looseness=-1

\noindent\textbf{Reviewing every chunk with a VLM is too slow on a real robot.}
Although effective in simulation~\cite{astra_embodied}, where query latency is ignored, the online agent fails in all 60 trials on our robot, as nearly every trial hits the 20\,min limit.
On count fruit, each query takes about 25\,s and the robot waits for every
reply, so it reaches only 4.0\% task progress within this time.
This failure comes from latency rather than the agent's reasoning:
given 30\,min on pack box, it reaches 42.7\% task progress
(\cref{sec:supp-online}).
\ours queries the VLM only offline, so no VLM latency enters the
control loop.\looseness=-1

\noindent\textbf{One-shot waypoint handoff leaves the policy unchecked.}
\ours-waypoint receives the same plan but passes it to the policy only
once per stage.
It nearly matches \ours on pack box, where choosing the targets is the
main difficulty, but succeeds in at most 4/15 trials on the other three
tasks.
After reaching the pre-grasp waypoint, the policy acts alone and can move toward a similar
distractor or an item that has already been counted, whereas the field
checks every chunk until the stage ends.\looseness=-1

\noindent\textbf{Higher success comes at little cost in execution time.}
In \cref{tab:time} (which omits puzzle, as baselines terminate early on most trials), \ours raises success on count fruit from 1/15 to 12/15 while matching the base policy's time per sub-goal (17.7\,s vs.\ 18.3\,s), showing that chunk evaluation adds negligible overhead; its trial time is longer (119.6\,s vs.\ 74.8\,s) only because it completes more items.
On pack earphone and pack box, \ours takes longer per sub-goal than the base policy and \ours-waypoint because it corrects deviations and retries failed grasps or insertions rather than failing early, raising task progress from 16.3--40.6\% to 82.8--83.3\%.
By contrast, the online agent takes 863.6--1733.3\,s per sub-goal and fails in all trials. These results answer \textbf{Q2}.\looseness=-1

\subsection{Effect of the Agent Model}
\label{sec:models}
To answer the agent side of \textbf{Q3}, we fix the policy and change only the
agent model on pack earphone (\cref{tab:models}).\looseness=-1

\noindent\textbf{A stronger agent model raises success.}
GPT-6-Astra and GPT-6.1-Sol reach 11/15 and 10/15, whereas no other model
exceeds 7/15.
Weaker models also fail to compile a valid field more often: GPT-6-Luna
fails in 6 of the 15 compilations (counting as 0\% progress).\looseness=-1

\begin{table}[t]
  \centering
    \caption{\textbf{Effect of the agent model.}
    Results on pack earphone with XR1 as the policy.
    Cost, compilation time, and input/output token counts are averaged per compilation, or per trial for the online agent in the top row.
    Qwen3.8-27B is self-hosted and incurs no API cost.}
  \label{tab:models}
  \footnotesize
  \setlength{\tabcolsep}{2.0pt}
  \begin{tabular}{@{}l cc ccc@{}}
    \toprule
    Model & SR & Prog. & Cost (\$) & Compile (s) & Tokens \\
    \midrule
    GPT-6-Astra (online) & 0/15  & 0.0  & 17.87 & --            & 12.91M/19.0k \\
    \midrule
    Qwen3.8-27B          & 7/15  & 59.4 & --    & 2629$\pm$810  & 8.20M/91.4k \\
    GPT-6-Luna           & 3/15  & 36.7 & 0.035 & 299$\pm$144   & 1.91M/15.7k \\
    Gemini-4-Argon       & 5/15  & 53.9 & 0.716 & 424$\pm$125   & 0.84M/14.8k \\
    Opus 5.5             & 7/15  & 76.7 & 4.000 & 574$\pm$147   & 1.80M/37.0k \\
    GPT-6.1-Sol          & 10/15 & 88.6 & 0.225 & 270$\pm$74    & 0.62M/4.9k \\
    GPT-6-Astra          & 11/15 & 82.8 & 1.437 & 339$\pm$55    & 0.67M/4.9k \\
    \bottomrule
  \end{tabular}
\end{table}

\noindent\textbf{A valid field is not enough.}
Gemini-4-Argon and Opus 5.5 compile a syntactically valid field every time, yet succeed in only 5/15 and 7/15 trials, as small errors in their goals or instance bindings are amplified during contact-rich execution.\looseness=-1

\noindent\textbf{Compilation cost is small and paid once per layout.}
With GPT-6-Astra, compiling the field once costs \$1.437 and 0.67M input tokens, an order of magnitude less than the \$17.87 and 12.91M tokens per trial of the online agent. GPT-6.1-Sol gives the best trade-off, with the highest progress at
\$0.225 and 270\,s per compilation, and the two strongest models use the fewest tokens, with about 90\% of their input served from cache.
The self-hosted Qwen3.8-27B reaches 7/15 with no API cost, showing that an open-weight model can already compile effective fields; although compilation takes 2629\,s, it finishes before the robot moves and can be accelerated with faster serving in future work.\looseness=-1

\subsection{Effect of the Policy}
\label{sec:policies}

\begin{figure}[t]
  \centering
  \includegraphics[width=\linewidth]{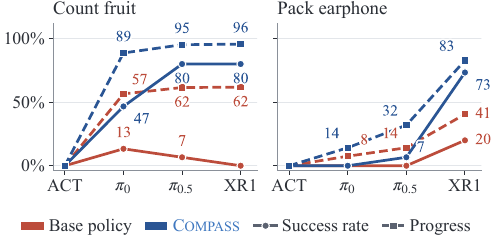}
    \caption{\textbf{Effect of the policy.}
    Base policy and \ours across four policies on count fruit and pack earphone, with the agent and the field fixed.
    Success rate is the share of 15 trials that succeed.}
  \label{fig:policies}
  \vspace{-8pt}
\end{figure}

To answer the policy side of \textbf{Q3}, we evaluate \ours with ACT~\cite{zhao2023act},
$\pi_0$~\cite{black2024pi0}, $\pi_{0.5}$~\cite{black2025pi05}, and
XR1~\cite{xiaomiroboticsteam2026xiaomirobotics1scalingvisionlanguageactionmodels}
on count fruit and pack earphone (\cref{fig:policies}), keeping the agent
and compiled fields fixed and training all policies on the same task demonstrations.\looseness=-1

\noindent\textbf{The field cannot compensate for missing motor skills.}
With ACT, \ours fails every trial on both tasks, mostly
through failed grasps or imprecise insertions.
Because the field corrects only pre-contact approach and leaves grasping and placing to the policy (\cref{sec:supp-control}), a policy
that cannot perform this contact also fails with \ours.\looseness=-1

\noindent\textbf{More precise contact requires a stronger policy.}
On count fruit, success rises from 7/15 with $\pi_0$ to 12/15 with
$\pi_{0.5}$ and XR1.
On pack earphone, where each small earbud must be placed into its case,
only XR1 succeeds in more than 1/15 trials, reaching 11/15.
We attribute this difference to XR1's extensive pretraining on the same robot embodiment and controller setup, rather than to the design of the policy itself.
Together with \cref{sec:models}, these results answer \textbf{Q3}: success
requires both a capable agent and a capable policy.\looseness=-1

\subsection{Effect of the Interface}
\label{sec:interface}

\begin{figure}[t]
  \centering
  \includegraphics[width=\linewidth]{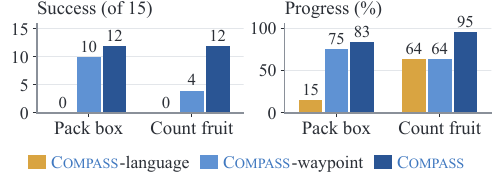}
    \caption{\textbf{Effect of the interface.}
    The same plan is passed to the policy as a language prompt in \ours-language, as waypoints in \ours-waypoint, or as a task field in \ours.}
  \label{fig:interfaces}
  \vspace{-8pt}
\end{figure}

To answer \textbf{Q4}, we pass the same plan to the same policy through three
interfaces (\cref{fig:interfaces}): a prompt in \ours-language, waypoints
in \ours-waypoint, and the task field in \ours.
We compare them on pack box and count fruit, whose training prompts name the objects.
\ours-language follows the common practice of language
subgoals~\cite{ahn2022saycan,shi2025hirobot}: it executes every policy
chunk unchanged and updates the prompt at every stage transition to name
what remains (\cref{sec:supp-baselines}).
It fails in all 15 trials on both tasks, with task progress remaining close to the base policy (15.1\% vs.\ 16.3\% on pack box and 63.8\% vs.\ 61.5\% on count fruit).
Waypoints raise success to 10/15 and 4/15, and the task field to 12/15
on both tasks.
These results answer \textbf{Q4}: language subgoals alone do not prevent the policy from drifting toward nearby distractors, whereas the task field enforces the plan throughout execution.\looseness=-1

\section{Conclusion}
\label{sec:conclusion}

We presented \ours, which connects agentic reasoning and a frozen visuomotor
policy through a task field compiled before execution.
A VLM agent makes the high-level decisions once and encodes them
as a stateful potential.
At run time, the field checks each policy chunk and leaves all contact to
the policy, so no VLM is queried in the control loop.
On four real-world bimanual tasks, \ours raises the average success rate
from 6.7\% to 76.7\%, outperforming online VLM guidance, language subgoals, and waypoint handoffs while scaling with both stronger agents and stronger policies.

\noindent\textbf{Limitations.}
The field is compiled from the initial layout, so large scene changes such as a displaced object require recompilation; triggering online recompilation upon unexpected disturbances is left to future work.
Errors in segmentation and monocular depth can shift compiled goals and bumps, which could be mitigated by more accurate depth estimation~\cite{yang2024depthanything,yang2024depthanythingv2,lin2025promptda,lin2025da3,ke2024marigold,piccinelli2024unidepth,bochkovskii2025depthpro,wang2025moge,wang2025vggt}
and robot geometry perception~\cite{liu2026mas}.
Finally, because the field guides pre-contact motion rather than adding new motor skills, each task requires a policy capable of the local contact.

\section*{LLM Usage Statement}
\noindent\textbf{Method and experimental evaluation.}
Vision-language models (VLMs) are a core component of \ours and the evaluated baselines (\cref{sec:models,tab:models}). Specifically, we use GPT-6-Astra (in the Codex CLI agent harness) as the default VLM agent for compiling task fields and for the VoxPoser and online agent baselines. In the agent model ablation (\cref{sec:models,tab:models}), we additionally evaluate GPT-6.1-Sol, GPT-6-Luna, Gemini-4-Argon, Opus 5.5, and Qwen3.8-27B.

\noindent\textbf{Manuscript polishing.}
Claude Code, OpenAI Codex, and Google Gemini were used to polish and refine the text and formatting across the manuscript and supplementary material. All text, experimental results, and citations were reviewed and verified by the authors, who take full responsibility for the contents of the paper.

\section*{Acknowledgment}
The SJTU team is partially supported by National Natural Science Foundation of China (62322603) and Shanghai Municipal Special Program for Basic Research on General AI Foundation Models (Grant No. 2025SHZDZX025D08).
{
    \small
    \bibliographystyle{ieeenat_fullname}
    \bibliography{main}
}

\clearpage
\maketitlesupplementary
\providecommand{\SO}{\mathrm{SO}(3)}

\section{Task Field Details}
\label{sec:supp-field}

During compilation, the agent selects the active terms in the stage potential $V_z$ \eqref{eq:field} and sizes the target tolerances ($\Sigma_k, \theta_{\text{tol}}$), clearance margins $r_j$, and policy workspaces $W_k$ from the target geometry and expected perception uncertainty (\cref{sec:compile}).
\Cref{tab:supp-field} summarizes the typical values chosen by the agent across our tasks.
Specifically, the orientation penalty $\psi(R, R_k^\ast)$ in \eqref{eq:field} is implemented as the normalized quadratic $\tfrac12 (\theta/\theta_{\text{tol}})^2$, where $\theta$ is the geodesic angle between $R$ and $R_k^\ast$ (or the angle between their approach axes when twist is free), and the gripper term $(g - g_k^\ast(p))^2$ is likewise normalized by its tolerance.

\noindent\textbf{Non-target bumps.}
Each repulsive bump $\phi_i(p)$ in \eqref{eq:field} depends on the normalized distance
$u_i = \|\Lambda_i^{-1}(p - c_i)\|$ to a non-target region
$i \in \mathcal{N}_z$, where $c_i$ is the region center and the
diagonal matrix $\Lambda_i$ holds its 3D half-extents (semi-axes) enlarged by a safety margin.
We use a compactly supported radial profile
\begin{equation}
  \phi_i(p) = w_i\, \beta_i(u_i), \qquad
  \beta_i(u) =
  \begin{cases}
    1 - 0.1\,u^2 & u \le 1, \\
    q_i(u) & 1 < u < \kappa_i, \\
    0 & u \ge \kappa_i ,
  \end{cases}
  \label{eq:bump}
\end{equation}
where the quintic polynomial $q_i$ decays smoothly from $0.9$ at the region
boundary to zero at the support radius $\kappa_i$, with coefficients chosen so
that $\phi_i$ is twice continuously differentiable.
We set $\kappa_i = \min\big(2.5,\, (1 + u_i(p_k^\ast))/2\big)$ so that each bump
vanishes before the stage goal $p_k^\ast$ and never shifts its minimum.
For regions whose geometry departs significantly from an ellipsoid, such as elongated produce or puzzle
pieces, we extrude the segmented 2D footprint to the object height and
apply the clearance barrier of \eqref{eq:field} to its signed distance.
Both forms vanish outside their support and sum into $\Phi_z = \sum_{i \in \mathcal{N}_z} \phi_i$.

\begin{table}[ht]
  \centering
  \caption{\textbf{Typical compiled field parameters.}
  Ranges of stage-specific parameters chosen by the agent across the four tasks.}
  \label{tab:supp-field}
  \small
  \setlength{\tabcolsep}{3pt}
  \begin{tabular}{@{}>{\raggedright\arraybackslash}p{0.47\linewidth}>{\raggedright\arraybackslash}p{0.49\linewidth}@{}}
    \toprule
    Parameter & Typical value \\
    \midrule
    Target tolerance $\Sigma_k$ (diagonal) & 4--8\,cm \\
    Clearance margin $r_j$                 & 0.8--1.2\,cm \\
    Margin of a non-target region          & 1.2--2.5\,cm \\
    Orientation weight $\lambda_R$         & 0.25--2 \\
    Orientation tolerance $\theta_{\text{tol}}$ & $10^\circ$--$45^\circ$ \\
    Policy workspace $W_k$                 & 6--8\,cm around the target \\
    \bottomrule
  \end{tabular}
\end{table}

\noindent\textbf{Transition predicates.}
During compilation, the agent assembles $E_k$ from lightweight proprioceptive and image checks and sets their thresholds for each stage, allowing $\delta$ to evaluate transitions at control rate without querying the VLM.
For grasping and placement stages, $E_k$ checks gripper closure under object contact followed by a short lift, or gripper opening within the compiled placement region.
Where visual verification is required, the agent attaches an appearance check on the cropped wrist or head view to confirm target presence or clearance and to reject grasped distractors.
When a predicate returns \emph{false} or a stage exceeds its timeout window, $\delta$ transitions to a recovery stage (\cref{sec:supp-exec}).

\section{Task Field Compilation Details}
\label{sec:supp-compile}

\subsection{Camera--Robot Calibration}
\label{sec:supp-calib}
Because the task field is compiled from the head-camera view but evaluated in the
base frame of each arm, we formulate a board-free calibration that maps image pixels to the 3D robot frame once per robot.
Consider $L \ge 1$ reference keypoints on the gripper with unknown offsets $\{b_l \in \mathbb{R}^3\}_{l=1}^L$ in the end-effector frame, moved across $N$ non-degenerate arm configurations $\{(p_n, R_n)\}_{n=1}^N$.
Because some keypoints may be self-occluded at certain poses, the user simply labels the visible subset of 2D keypoints $\{u_{nl}\}_{(n,l) \in \mathcal{O}}$ in the head images, and the agent executes the rest of the calibration pipeline automatically.
With the camera intrinsics $K$ fixed, the agent jointly estimates, separately for each arm, the camera
extrinsics $(R_c, t_c) \in \mathrm{SE}(3)$ and keypoint offsets $\{b_l\}$ ($6 + 3L$ degrees of freedom in total) from the $2|\mathcal{O}| \ge 6 + 3L$ pixel constraints by minimizing the robust reprojection error
\begin{equation}
  \min_{R_c,\, t_c,\, \{b_l\}} \; \sum_{(n,l) \in \mathcal{O}} \rho\Big( \big\| \Pi_K\big(R_c (p_n + R_n b_l) + t_c\big) - u_{nl} \big\|^2 \Big),
  \label{eq:calib}
\end{equation}
where $\Pi_K$ denotes perspective projection and $\rho$ is a Huber loss.
The agent initializes the optimization via PnP and validates the solution by leave-one-pose-out cross-validation, fitting on $N - 1$ poses and evaluating on the held-out pose.
In our setups, we label $L = 4$ gripper landmarks (two fingertips and two screws) across $N \approx 15\text{--}16$ poses per arm, achieving held-out root-mean-square reprojection errors on the left and right arms of 2.47\,px and 2.42\,px on the RealSense D405 setup ($640\times480$, 31 valid poses) and 2.68\,px and 2.69\,px on the rectified fisheye setup ($960\times720$, 32 valid poses).

\subsection{Perception}
\label{sec:supp-perception}

During compilation, the agent segments task-relevant objects and parts in the head image using SAM~3~\cite{carion2025sam3} with positive and negative point prompts, and predicts dense depth using Depth Anything~3~\cite{lin2025da3}.
Both models run on a remote GPU server and are queried only once before execution.
To recover metric depth from the relative monocular predictions, the agent anchors the predicted depth map of each scene to a reference support plane and back-projects each segmented 2D mask through the calibrated extrinsics $(R_c, t_c)$ (\cref{sec:supp-calib}) into a 3D bounding extent in the robot frame.

\subsection{Agent and Compilation Pipeline}
\label{sec:supp-agent}
By default, the VLM agent runs GPT-6-Astra with medium reasoning effort in the Codex CLI; all agent models evaluated in \cref{tab:models} share the same CLI harness, skill documents, and tool interfaces.
Before compilation begins, both arms move to a home pose so the head camera captures an unobstructed workspace view $o_0$.
The compilation time reported in \cref{tab:models} measures the complete wall-clock duration from this initialization through all model queries, perception calls, and verification steps.

\noindent\textbf{Skill hierarchy.}
Once per task, the agent constructs a reusable skill document from the task description, a few demonstrations, and brief autonomous exploration (\cref{sec:compile}).
The document is organized into four modular layers:
(1) core interface rules specifying which constraints belong in the task field versus the visuomotor policy;
(2) the robot embodiment and calibrated camera-to-robot transforms (\cref{sec:supp-calib});
(3) reusable task-family templates (e.g., sequential packing) defining standard stage transitions, candidate potential terms, and predicate types; and
(4) task-specific semantics, including the policy's training-prompt vocabulary, ordering or quota rules, and instance-binding logic.
Tasks within the same family share the first three layers and differ only in the fourth.

\noindent\textbf{Tool interface and visual verification.}
To execute the four compilation steps of \cref{sec:compile} on a new layout, the agent interacts with the workspace through three modular tools:
a perception tool that takes positive and negative point prompts on the head image and returns 2D masks and robot-frame 3D bounding extents (\cref{sec:supp-perception});
a field-synthesis tool that validates the compiled stage ledger $m_0$, spatial goals $(p_k^\ast, R_k^\ast)$, selective potential terms in $V_z$, policy workspaces $W_k$, and predicates $E_k$ (\cref{tab:supp-field});
and a verification renderer that projects the compiled 3D extents, stage goals $(p_k^\ast, R_k^\ast)$, policy workspaces $W_k$, and non-target bumps $\phi_i$ back onto the head-camera image.
Inspecting this composite overlay allows the agent to detect segmentation drift, misbound instances, or colliding approach axes and to re-prompt perception or adjust field parameters before robot execution starts.

\section{Field-Guided Execution Details}
\label{sec:supp-exec}

\Cref{alg:execution} summarizes the run-time evaluation of each action chunk for one arm; in bimanual stages, both arms are evaluated independently against their respective potentials and share the field-state update $\delta$.

\begin{algorithm}[ht]
\caption{One chunk of field-guided execution for one arm.}
\label{alg:execution}
\begin{algorithmic}[1]
\Require compiled field $\mathcal{F}$, field state $z=(k, m)$, observation $o_t$, measured state $s_t$
\State $A_t \gets s_t \oplus \pi(s_t, o_t, \ell_k)$
\If{no term in $V_z$ is active for this arm}
  \State execute $A_t$
\ElsIf{$s_t \in W_k$}
  \State repair small exits of $A_t$ from $W_k$ (or re-query $\pi$ on larger ones), then execute
\ElsIf{$A_t$ satisfies \eqref{eq:admissible}}
  \State execute $A_t$
\Else
  \State execute bounded gradient correction \eqref{eq:descent} (with detour goal if blocked)
\EndIf
\State observe post-chunk state and images $(s_{t'}, o_{t'})$
\State $z \gets \delta\big(z, E_k(s_{t'}, o_{t'})\big)$
\end{algorithmic}
\end{algorithm}

\noindent\textbf{Detours around local minima.}
When the straight segment from $p_t$ to $p_k^\ast$ intersects a non-target region or gradient descent stalls (\cref{sec:execution}), the field selects a collision-free temporary waypoint just outside the support radius $\kappa_i$ of the blocking region (above or laterally around its boundary) that minimizes distance to $p_k^\ast$, and switches back to $p_k^\ast$ once the direct line of sight is clear.

\noindent\textbf{Recovery stages.}
When $E_k$ detects an empty or mismatched grasp or a stage timeout (\cref{sec:supp-field}), $\delta$ transitions to a recovery stage that opens the gripper, guides the arm back to a pre-grasp clearance pose above the target, and hands control back to the policy upon re-entering $W_k$, pausing execution if the stage's retry budget is exhausted.

\noindent\textbf{Run-time execution settings.}
\Cref{tab:supp-exec} summarizes the run-time settings of field-guided execution.
At 30\,Hz, each policy query predicts an $H=30$-frame chunk, of which 15--30 frames are executed before the next query.
When constructing the task skill (\cref{sec:compile}), the agent sets the per-chunk correction budgets and potential tolerance $\tau$ to match the task's workspace scale and policy stride, while the line-search and path-discretization constants remain fixed across all tasks without manual tuning.

\begin{table}[ht]
  \centering
  \caption{\textbf{Run-time execution settings.}
    Correction budgets and the potential tolerance $\tau$ are set by the agent in each task skill, while numerical solver and discretization constants are fixed across all tasks.}
  \label{tab:supp-exec}
  \small
  \setlength{\tabcolsep}{4pt}
  \begin{tabular}{@{}l l@{}}
    \toprule
    Setting & Typical value \\
    \midrule
    Policy chunk horizon $H$                & 30 frames (1\,s at 30\,Hz) \\
    Executed frames per chunk               & 15--30 \\
    Correction horizon                      & up to 30 frames \\
    Translation budget per correction       & 5--20\,cm \\
    Rotation budget per correction          & $10^\circ$--$20^\circ$ ($\le 1^\circ$/frame) \\
    Backtracking line-search steps          & up to 24 \\
    Potential tolerance $\tau$ in \eqref{eq:admissible} & $\le 0.02$ \\
    Inactive bump threshold $\epsilon$      & 0.05 \\
    Path discretization step                & 2\,mm, $2^\circ$, 0.05 gripper \\
    \bottomrule
  \end{tabular}
\end{table}

\begin{figure*}[t]
  \centering
  \includegraphics[width=\textwidth]{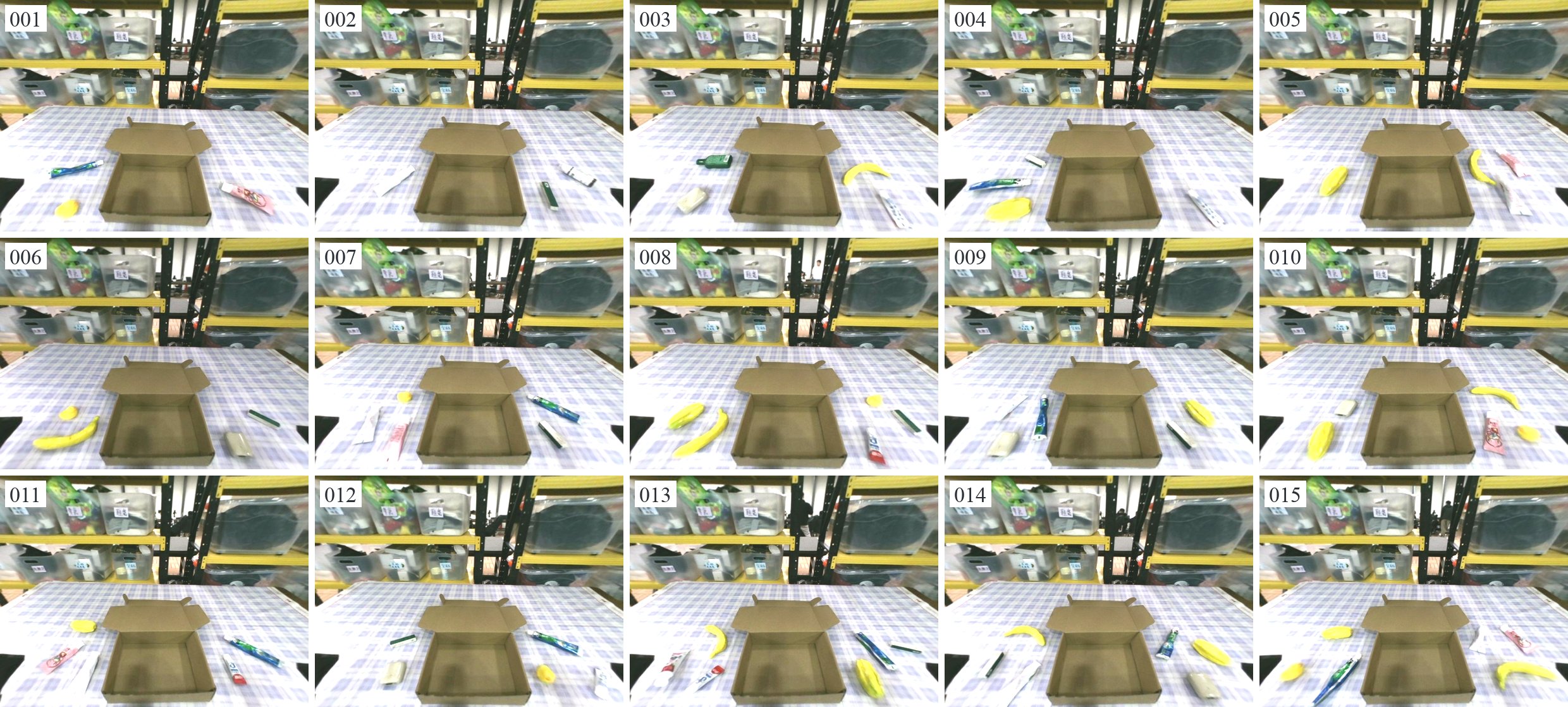}
  \caption{\textbf{Test layouts of pack box.}
  All 15 evaluation layouts shown in the head-camera view.}
  \label{fig:supp-layouts-box}
\end{figure*}

\begin{figure*}[t]
  \centering
  \includegraphics[width=\textwidth]{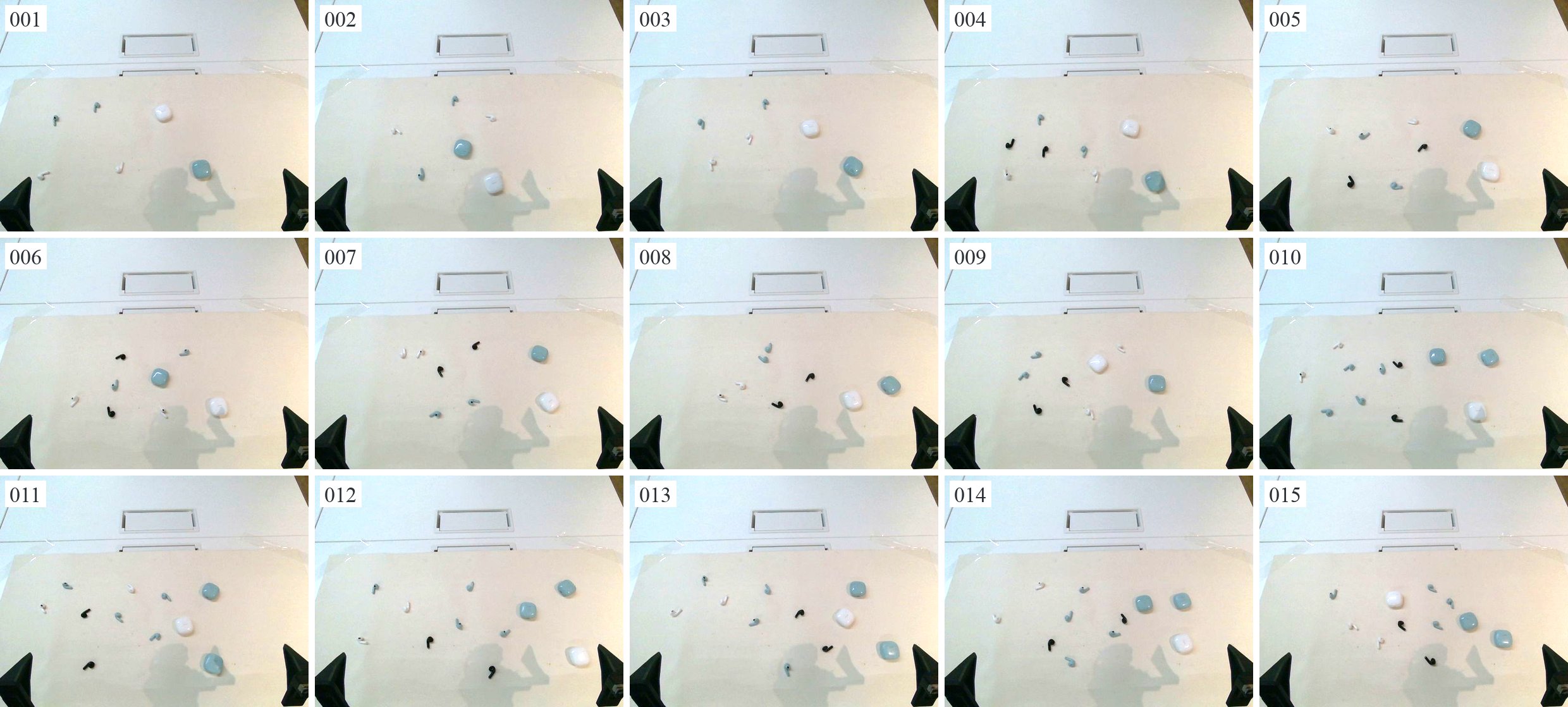}
  \caption{\textbf{Test layouts of pack earphone.}
  All 15 evaluation layouts shown in the head-camera view.}
  \label{fig:supp-layouts-earphone}
\end{figure*}

\begin{figure*}[t]
  \centering
  \includegraphics[width=\textwidth]{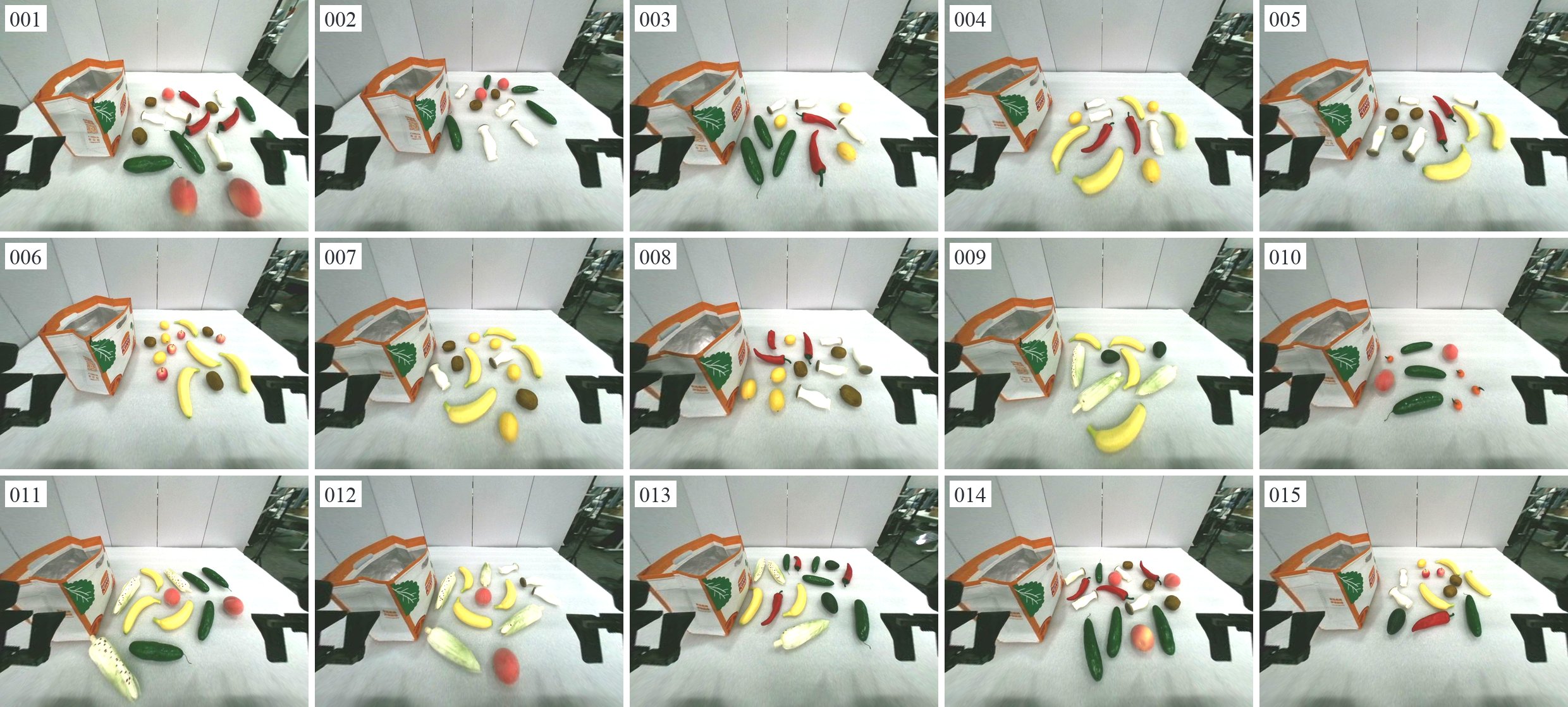}
  \caption{\textbf{Test layouts of count fruit.}
  All 15 evaluation layouts shown in the head-camera view.}
  \label{fig:supp-layouts-fruit}
\end{figure*}

\begin{figure*}[t]
  \centering
  \includegraphics[width=\textwidth]{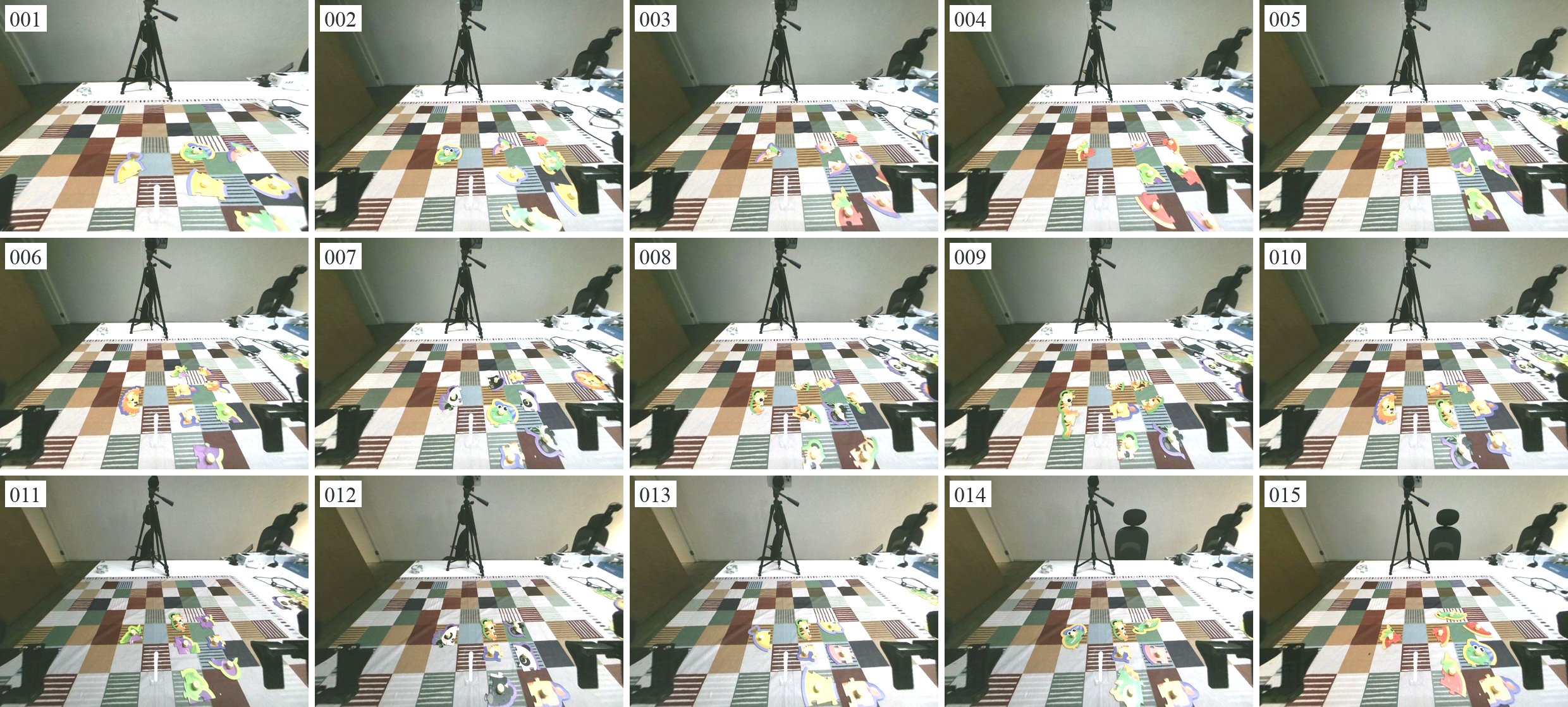}
  \caption{\textbf{Test layouts of puzzle.}
  All 15 evaluation layouts shown in the head-camera view.}
  \label{fig:supp-layouts-puzzle}
\end{figure*}

\section{Experimental Setup Details}
\label{sec:supp-setup}

\subsection{Robots and Policies}
\label{sec:impl}

\noindent\textbf{Robot platform.}
All experiments are conducted on bimanual platforms equipped with two 6-DoF ARX X5 arms, parallel-jaw grippers, one head camera, and two wrist-mounted cameras (\cref{sec:setup}).
Each arm is controlled in its own base frame, calibrated to the head camera via \cref{sec:supp-calib}.
The pack earphone setup uses three Intel RealSense D405 RGB cameras ($640\times480$), while pack box, count fruit, and puzzle use fisheye RGB cameras ($960\times720$, undistorted before calibration and compilation).
Only RGB streams are used across all tasks; all 3D geometry is derived from monocular depth estimation.
High-level chunk evaluation and policy inference run at a 30\,Hz action rate, tracked by a 200\,Hz low-level joint controller.

\noindent\textbf{Base policies.}
We use XR1~\cite{xiaomiroboticsteam2026xiaomirobotics1scalingvisionlanguageactionmodels} for pack earphone and puzzle and $\pi_{0.5}$~\cite{black2025pi05} for pack box and count fruit, each served on a single RTX 4090 GPU (together with the additional policy backbones evaluated in \cref{fig:policies}).
At each query, the policy maps the three RGB views $o_t$, measured arm states $s_t$, and stage prompt $\ell_k$ to $H=30$ relative actions per arm, which compose with $s_t$ into absolute action chunks $A_t$ in the arm base frames.
All policies are post-trained on teleoperated task demonstrations: 10.0\,h for pack earphone, 10.2\,h for pack box, 4.6\,h for count fruit, and 10.6\,h of demonstrations supplemented with 2.7\,h of DAgger rollouts for puzzle.
During execution, the policy receives prompts matching its training distribution (e.g., \texttt{pack earphone} or object names from the training vocabulary) rather than the raw free-form user instruction.

\subsection{Tasks}
\label{sec:supp-tasks}

\noindent\textbf{Pack box.}
Each trial presents a free-form instruction naming 3--6 objects to be packed in a specific sequence (70 target objects across the 15 layouts, drawn from 11 unseen objects in novel target combinations and orders), such as \emph{``Pick up the teal toothpaste tube, the pink toothpaste tube, and the yellow rubber duck, in that order, and place each one into the box.''}
The agent resolves the instruction into an ordered target sequence and binds each named item to its 3D instance and corresponding training-vocabulary prompt.
A trial succeeds only if all requested objects are placed into the box in the exact specified order.

\noindent\textbf{Pack earphone.}
Layouts 1--9 contain two seen earphone sets and layouts 10--15 contain three (36 sets in total across seen and unseen color combinations, including duplicate same-color sets), with several layouts also including extra black distractor earbuds.
Each set comprises a charging case and a pair of left and right earbuds.
The right arm picks up and holds the case open while the left arm sequentially picks and inserts the two matching earbuds, after which the case is closed and placed back on the table.
When multiple sets share the same color, the agent must bind each earbud to a distinct set instance and protect already packed sets as non-target regions in subsequent stages.
A trial succeeds when every set is packed with its matching earbuds, closed, and placed on the table.

\noindent\textbf{Count fruit.}
The left arm holds a deep bag open while the right arm packs seen fruit and vegetable categories from the 10-item training vocabulary to satisfy novel category quotas and counts such as \emph{``a couple of each''} or \emph{``three bananas and two lemons''} (5--10 items per layout, 120 in total).
Because placed items drop out of sight inside the bag, a memoryless policy cannot infer how many items have already been packed from current images alone.
A trial succeeds if the bag contains the exact requested count of each item category.

\noindent\textbf{Puzzle.}
Each layout places 2--5 seen handled jigsaw pieces (49 targets in total across the 15 layouts) on the table in unseen configurations together with 2--4 distractor pieces from other puzzles (48 in total).
While the left arm stabilizes the anchor piece, the agent infers which pieces belong to the puzzle, their assembly order, and their target poses, guiding the right arm to grasp and fit each piece into the assembly.
A trial succeeds when all target pieces of the puzzle are fitted.

\Cref{fig:supp-layouts-box,fig:supp-layouts-earphone,fig:supp-layouts-fruit,fig:supp-layouts-puzzle} show all 15 test layouts for each of the four tasks.

\subsection{Evaluation Protocol}
\label{sec:supp-eval}

\noindent\textbf{Sub-goals and task progress.}
On pack box and count fruit, each requested item counts as one sub-goal: 1.0 credit when picked and placed into the container, and 0.5 partial credit when grasped without successful placement.
On pack earphone, each earphone set counts as one sub-goal, divided equally into four steps of 0.25 credit: picking and opening the case, inserting the first earbud, inserting the second earbud, and closing and placing the case.
On puzzle, each fitted target piece (excluding the pre-held anchor piece) counts as one sub-goal.
Task progress for a trial is the number of completed sub-goals (including partial credit) divided by the total number of sub-goals in that layout, averaged uniformly across the 15 layouts.

\noindent\textbf{Stopping rules.}
Every trial terminates upon completion, after 20\,min, or immediately upon an unrecoverable task error: inserting an earbud into a mismatched case, packing an extra or unrequested item into the bag, placing an object into the box out of the requested order, or bringing a distractor piece to the puzzle assembly.

\noindent\textbf{Execution time.}
Trial execution time measures active robot execution and excludes offline field compilation; for the online agent, it includes the run-time wait for VLM queries during execution.
Trials that reach the 20\,min limit or stall until timeout are recorded as 1200\,s, whereas trials that succeed or terminate early at an unrecoverable error record their elapsed execution time.
In \cref{tab:time}, \emph{Time / trial} averages this duration over all 15 trials, and \emph{Time / sub-goal} divides each trial's duration by the number of completed sub-goals (including partial credit), averaged over trials with nonzero progress (so a timed-out trial with only 0.5 partial credit contributes 2400\,s per sub-goal).
When per-trial controller timestamps are not logged by a baseline, execution durations are measured from the synchronized trial recordings.

\noindent\textbf{Compute and cost.}
\ours requires no additional policy training.
Post-training all visuomotor policies across the four tasks and policy ablations required approximately 6.0k H200-equivalent GPU-hours.
All VLM and agent API queries across field compilation, online agent baselines, VoxPoser, model ablations, and development totaled approximately \$1.2k at list prices, and the 585 real-robot evaluation trials required roughly 200 operator hours including scene setup and resets.

\subsection{Policy Prompts}
\label{sec:supp-prompts}

\noindent\textbf{Training prompts.}
Each demonstration is labeled with a single episode-level prompt that stays
fixed for the whole episode (\cref{tab:supp-prompts}).
The pack earphone and puzzle policies are trained with one task token each,
so their prompt carries no information about the scene.
The pack box and count fruit policies are trained with prompts that list the
objects in the order they were packed in the demonstration, using a fixed
template and a closed vocabulary of short object names:
90 object names for pack box, such as \texttt{green tube} or
\texttt{panda plush toy}, and 10 fruit and vegetable names for count fruit.

\begin{table*}[t]
  \centering
  \caption{\textbf{Prompts used to post-train the base policies.}
  Each demonstration has one prompt for the whole episode;
  $\langle o_i\rangle$ is a name from the task's training vocabulary.}
  \label{tab:supp-prompts}
  \small
  \begin{tabular}{@{}l l l@{}}
    \toprule
    Task & Training prompt & Vocabulary \\
    \midrule
    Pack earphone & \texttt{pack earphone} & task token \\
    Puzzle & \texttt{puzzle\_piece\_fit\_lh.} & task token \\
    Pack box & \texttt{Put $\langle o_1\rangle$, \ldots{} and $\langle o_n\rangle$ in the box.} & 90 object names \\
    Count fruit & \texttt{Pick $\langle o_1\rangle$, \ldots, $\langle o_n\rangle$ in this order and pack them into the bag.} & 10 item names \\
    \bottomrule
  \end{tabular}
\end{table*}

\noindent\textbf{Test-time prompts.}
On pack earphone and puzzle, every method sends the task token.
On pack box and count fruit, the targets named in the instruction are written
into the training template.
Many test objects are not in the training vocabulary. Each is described by
the closest training name in category and appearance; for example, a teal
toothpaste tube becomes \texttt{green tube}.
When no training name is close, the object keeps a short appearance
description.
This mapping changes only the text sent to the policy.

\begin{table}[t]
  \centering
  \caption{\textbf{Policy prompts at test time on pack box.}}
  \label{tab:supp-prompt-examples}
  \small
  \begin{tabularx}{\columnwidth}{@{}l X@{}}
    \toprule
    Method & Pack box \\
    \midrule
    Base policy & ground-truth ordered list, fixed for the trial \\
    \ours-language & list of remaining targets, updated at each stage transition \\
    \ours / \ours-waypoint & compiled ordered list, fixed for the trial \\
    \midrule
    Example & \emph{``Put green tube, pink tube and yellow plush toy in the box.''} \\
    \bottomrule
  \end{tabularx}
\end{table}

\noindent\textbf{Prompts in \ours.}
During compilation, the agent writes the stage prompts $\ell_k$
(\cref{sec:method}) in the training template and vocabulary.
The prompt therefore matches the training distribution in form, while its
object list can be a combination of training names never seen in any single
demonstration.
For example, on pack box (\cref{tab:supp-prompt-examples}), the prompt lists all targets and does not change across stages,
so that the policy sees the same prompt type as in training. Which object the
arm reaches in each stage is set by the field, not by the prompt.

\noindent\textbf{Prompts in \ours-language.}
\ours-language receives the same compiled targets, order, instance bindings,
and stage transitions as \ours, but its only channel to the policy is the
prompt.
The prompt is therefore updated at every stage transition in the training
template of each task to list the targets that remain.
For example, on pack box, it starts as the episode prompt of \ours and drops the object
whose stage has just ended, so it always lists the remaining targets in
order: \emph{``Put green tube, pink tube and yellow plush toy in the box.''},
then \emph{``Put pink tube and yellow plush toy in the box.''}, and finally
\emph{``Put yellow plush toy in the box.''}
The prompt changes only at stage transitions, which are
detected exactly as in \ours, so the policy is told what to do next at the
same moments as in \ours.

\section{Baselines}
\label{sec:supp-baselines}

\noindent\textbf{Base policy.}
The base policy executes the frozen visuomotor policy without agent compilation or field guidance, conditioned on the task's training-distribution prompt format.
On pack earphone and puzzle, where the policy is trained with a single task token (\texttt{pack earphone} and \texttt{puzzle\_piece\_fit\_lh.}), it receives that fixed prompt.
On pack box and count fruit, we provide the ground-truth ordered target list formatted in the training prompt template (for example, \emph{``Put green tube, pink tube and yellow plush toy in the box.''}), mapping any out-of-vocabulary object description to its closest training label.
Thus, on pack box and count fruit, the base policy is directly given the resolved target sequence and only needs to execute and track it.

\noindent\textbf{VoxPoser.}
Our re-implementation of VoxPoser~\cite{huang2023voxposer} uses the same VLM (GPT-6-Astra) and 3D perception outputs (SAM~3 and Depth Anything~3) as \ours.
The VLM synthesizes 3D affordance and avoidance value maps discretized at 4\,mm voxel resolution, and an A* motion planner generates end-effector trajectories on the 26-connected voxel grid with automated gripper triggers at target regions.

\noindent\textbf{Online agent.}
We adapt the hybrid online control paradigm of~\cite{astra_embodied}, in which GPT-6-Astra is queried at every chunk step to inspect the current camera views and policy proposal, choosing either to execute a prefix of the proposed policy chunk or to override it with end-effector waypoints.
For a fair comparison on real hardware, we initialize the online agent with the same perception, instance bindings, stage ledger, and transition predicates $E_k$ as \ours, using the same medium reasoning effort, while replacing offline field evaluation with online VLM queries at each chunk.
Unlike simulation benchmarks that pause physics during VLM inference, our real-robot evaluation includes query latency within the 20\,min trial limit.

\noindent\textbf{\ours-waypoint.}
This variant retains the exact compiled stages, instance bindings, language prompts $\ell_k$, and transition predicates $E_k$ of \ours, but replaces continuous field guidance with a one-shot waypoint handoff per stage.
A motion planner moves the active gripper to the compiled stage waypoint (within 2.5\,cm of the target) and hands full control to the policy, which then executes all subsequent chunks unchecked until $E_k$ triggers the next stage transition.

\noindent\textbf{\ours-language.}
This variant retains the compiled target sequence, quotas, and stage transitions of \ours, but removes both spatial waypoints and the task field.
Both arms are controlled solely by the visuomotor policy throughout the trial, and at each stage transition the language prompt $\ell_k$ is updated in the task's training template to list only the remaining targets to be packed.
Because the policies for pack earphone and puzzle are trained with a single fixed task prompt rather than object-conditioned language instructions, \ours-language is evaluated on pack box and count fruit.

\section{Additional Results}
\label{sec:supp-results}

\subsection{Control Sharing}
\label{sec:supp-control}

\begin{figure}[t]
  \centering
  \includegraphics[width=\linewidth]{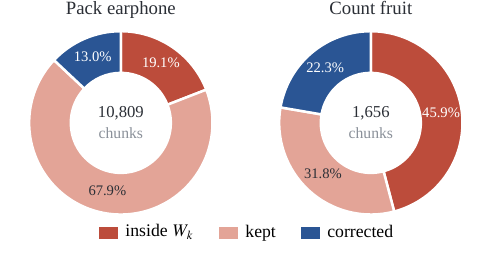}
  \caption{\textbf{Chunk execution breakdown.}
  Share (\%) of action chunks on pack earphone and count fruit across the 15 trials of \ours that are executed by the policy inside $W_k$, kept unchanged outside $W_k$, or replaced by a field correction.
  Total chunk counts are shown in the center.}
  \label{fig:supp-control-share}
\end{figure}

\Cref{fig:supp-control-share} breaks down how all action chunks across the 15 trials are executed on pack earphone and count fruit.
Field corrections replace only 13.0\% and 22.3\% of the chunks, respectively, while the remaining 87.0\% and 77.7\% are executed as proposed by the policy---inside $W_k$, in unconstrained stages where no term in $V_z$ is active, or as admissible chunks outside $W_k$.
On pack earphone, 865 of the 1,407 corrected chunks occur while the arm approaches a target earbud or case and the remainder mainly when the policy attempts to exit $W_k$, whereas none of the 5,030 earbud-insertion or case-manipulation chunks is modified.
Similarly, on count fruit, no chunk is modified while the policy grasps inside $W_k$, lifts an item, or places it into the bag.
Thus, the field steers target selection and approach while leaving contact-rich manipulation entirely to the policy.

\subsection{Online Agent with an Extended Time Horizon}
\label{sec:supp-online}

To separate the effect of VLM query latency from high-level reasoning quality, we allowed the online agent to run for up to 30\,min per trial on pack box (versus the standard 20\,min limit in \cref{sec:setup}).
When evaluated at the 20\,min mark from timestamped placement logs (\cref{fig:main,tab:time}), the online agent achieves 28.4\% task progress and 0/15 successes.
Extending the horizon to 30\,min increases its progress to 42.7\% and yields 1/15 successes, requiring an average of 947.7\,s per sub-goal compared with 38.1\,s for \ours.
This result indicates that the online agent is bottlenecked not by visual reasoning, but by the cumulative latency of querying a large VLM inside the closed-loop control cycle.

\subsection{Compilation Time Breakdown}
\label{sec:supp-compile-time}

\begin{figure}[t]
  \centering
  \includegraphics[width=\linewidth]{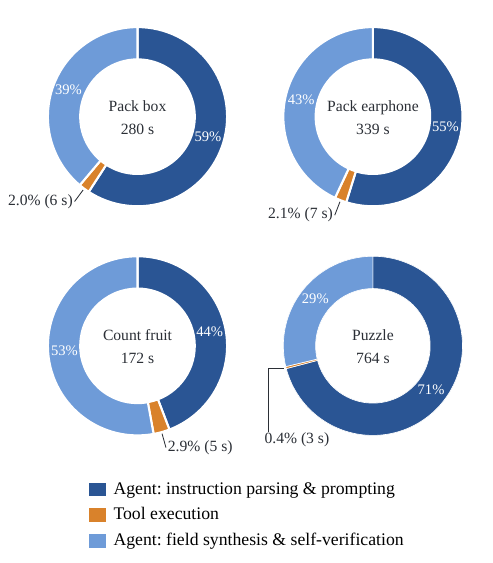}
  \caption{\textbf{Compilation time breakdown by task.}
  Share of compilation time across the four tasks spent on instruction parsing and point prompting, tool execution, and field synthesis and self-verification.
  Mean compilation time per layout is shown in the center.}
  \label{fig:supp-compile-time}
\end{figure}

\begin{table}[t]
  \centering
  \caption{\textbf{Compilation time by agent model.}
  Mean duration in seconds and share in \% on pack earphone across the 15 trials in \cref{tab:models}, grouped into the same three categories as \cref{fig:supp-compile-time}. Percentages may not sum to 100\% due to rounding.}
  \label{tab:supp-compile-models}
  \footnotesize
  \setlength{\tabcolsep}{0pt}
  \begin{tabular*}{\columnwidth}{@{\extracolsep{\fill}}l c ccc@{}}
    \toprule
    Model & Total (s) & \begin{tabular}[b]{@{}c@{}}Parsing \&\\[-1pt] prompting\end{tabular} & \begin{tabular}[b]{@{}c@{}}Tool\\[-1pt] execution\end{tabular} & \begin{tabular}[b]{@{}c@{}}Synthesis \&\\[-1pt] verification\end{tabular} \\
    \midrule
    Qwen3.8-27B    & 2629$\pm$810 & 2195 (83\%) & 6.7 (0.3\%) & 427 (16\%) \\
    GPT-6-Luna     & 299$\pm$144  & 164 (55\%)  & 8.7 (2.9\%) & 126 (42\%) \\
    Gemini-4-Argon & 424$\pm$125  & 325 (77\%)  & 6.5 (1.5\%) & 93 (22\%)  \\
    Opus 5.5       & 574$\pm$147  & 461 (80\%)  & 6.7 (1.2\%) & 106 (19\%) \\
    GPT-6.1-Sol    & 270$\pm$74   & 150 (56\%)  & 8.8 (3.3\%) & 111 (41\%) \\
    GPT-6-Astra    & 339$\pm$55   & 187 (55\%)  & 7.2 (2.1\%) & 145 (43\%) \\
    \bottomrule
  \end{tabular*}
\end{table}

\Cref{fig:supp-compile-time} breaks down the wall-clock compilation time of \ours (with GPT-6-Astra) across the four tasks into three parts corresponding to the workflow in \cref{sec:compile,sec:supp-agent}:
(1) agent reasoning for instruction parsing and point-prompt generation;
(2) tool execution, covering the initial home motion and camera capture, remote SAM~3 and Depth Anything~3 inference, and field synthesis and rendering; and
(3) agent reasoning for spatial goal specification, field parameterization, and visual self-verification.
Across all four tasks, tool execution accounts for only 3--7\,s per layout (0.4--2.9\% of the total time): robot initialization and capture take up to 2.3\,s, remote 3D perception takes 0.4--1.1\,s, and field construction and overlay rendering take 0.3--5.4\,s.
The remaining 97.1--99.6\% is spent in VLM reasoning, split between instruction parsing and prompting (44--71\%) and field synthesis and verification (29--53\%).
Puzzle takes the longest to compile (764\,s on average) because the agent reasons over fine-grained jigsaw contours and candidate mating orientations before placing point prompts.
Because perception and field construction themselves take only a few seconds, compilation time is governed almost entirely by VLM inference speed.

\Cref{tab:supp-compile-models} breaks down compilation time across the six VLM agent models evaluated on pack earphone (\cref{tab:models}) using the same three categories as \cref{fig:supp-compile-time}.
Across all models, tool execution remains nearly constant at 6.5--8.8\,s (0.3--3.3\%), while agent reasoning for field synthesis and self-verification takes 93--145\,s for all frontier API models.
The variation in total compilation time comes mainly from instruction parsing and point prompting: GPT-6.1-Sol, GPT-6-Luna, and GPT-6-Astra resolve the earphone sets and point prompts in 150--187\,s (55--56\%), whereas Gemini-4-Argon, Opus 5.5, and Qwen3.8-27B spend 325--2195\,s (77--83\%) inspecting cropped regions before calling perception.

\end{document}